\documentclass[letterpaper, 10 pt, conference]{ieeeconf}  

\IEEEoverridecommandlockouts                              

\usepackage{graphicx} 
\usepackage{amsmath} 
\usepackage{amssymb}  
\usepackage{url}
\usepackage{svg}
\usepackage{amsmath}
\usepackage{dsfont}
\usepackage{hyperref}
\usepackage{csquotes}

\title{\LARGE \bf Underwater Multi-Robot Convoying using\\
Visual Tracking by Detection}

\author{Florian Shkurti$^{1}$, Wei-Di Chang$^{1}$, Peter Henderson$^{1}$, Md Jahidul Islam$^{2}$, Juan Camilo Gamboa Higuera$^{1}$,\\
Jimmy Li$^{1}$, Travis Manderson$^{1}$, Anqi Xu$^{3}$, Gregory Dudek$^{1}$, Junaed Sattar$^{2}$
\thanks{$^{1}$Centre for Intelligent Machines and School of Computer Science, McGill University
        {\tt\footnotesize \{florian,wchang,petehend,gamboa,jimmyli,travism, \
		dudek\}@cim.mcgill.ca}}%
\thanks{$^{2}$Interactive Robotics and Vision Laboratory, Department of Computer Science and Engineering, University of Minnesota- Twin Cities
        {\tt\footnotesize \{islam034,junaed\}@umn.edu}}%
\thanks{$^{3}$Element AI {\tt\footnotesize ax@elementai.com} \emph{(work conducted as a post-doctoral researcher at McGill University with funding from NSERC)}}%
}

\begin{document}

\maketitle
\thispagestyle{empty}
\pagestyle{empty}

\begin{abstract}

%
%
%
%
%
%
%
%
%

We present a robust multi-robot convoying approach that relies on visual detection of the leading agent, thus enabling target following in unstructured 3-D environments.
Our method is based on the idea of \emph{tracking-by-detection}, which interleaves efficient model-based object detection with temporal filtering of image-based bounding box estimation.
This approach has the important advantage of mitigating tracking drift (i.e. drifting away from the target object), which is a common symptom of model-free trackers and is detrimental to sustained convoying in practice.
To illustrate our solution, we collected extensive footage of an underwater robot in ocean settings, and hand-annotated its location in each frame. Based on this dataset, we present an empirical comparison of multiple tracker variants, including the use of several convolutional neural networks, both with and without recurrent connections, as well as frequency-based model-free trackers.
We also demonstrate the practicality of this tracking-by-detection strategy in real-world scenarios by successfully controlling a legged underwater robot in five degrees of freedom to follow another robot's independent motion.
\end{abstract}

\section{INTRODUCTION}

Vision-based tracking solutions have been applied to robot convoying in a variety of contexts, including terrestrial driving~\cite{schneiderman1995vision,fries2014monocular}, on-rails maintenance vehicles~\cite{maire2007vision}, and unmanned aerial vehicles~\cite{lugo2013following}. Our work demonstrates robust tracking and detection in challenging underwater settings which are employed successfully for underwater convoying in ocean experiments.
This is achieved through \emph{tracking-by-detection}, which combines target detection and temporally filtered image-based position estimation. Our solution is built upon several autonomous systems for enabling underwater tasks for a hexapod robot~\cite{sattar2008enabling,sattar2009robust,sattar2009boosting,girdhar2014exploring,meger2015learning}, as well as recent advances in real-time deep learning-based object detection frameworks~\cite{redmon2016yolo9000, renNIPS15fasterrcnn}.

In the underwater realm, convoying tasks face great practical difficulties due to highly varied lighting conditions, external forces, and hard-to-model currents on the robot. While previous work in terrestrial and aerial systems used fiducial markers on the targets to aid tracking, we chose a more general tracking-by-detection approach that is trained solely on the natural appearance of the object/robot of interest.
\begin{figure}[ht]
\centering
\includegraphics[width=1.0\linewidth]{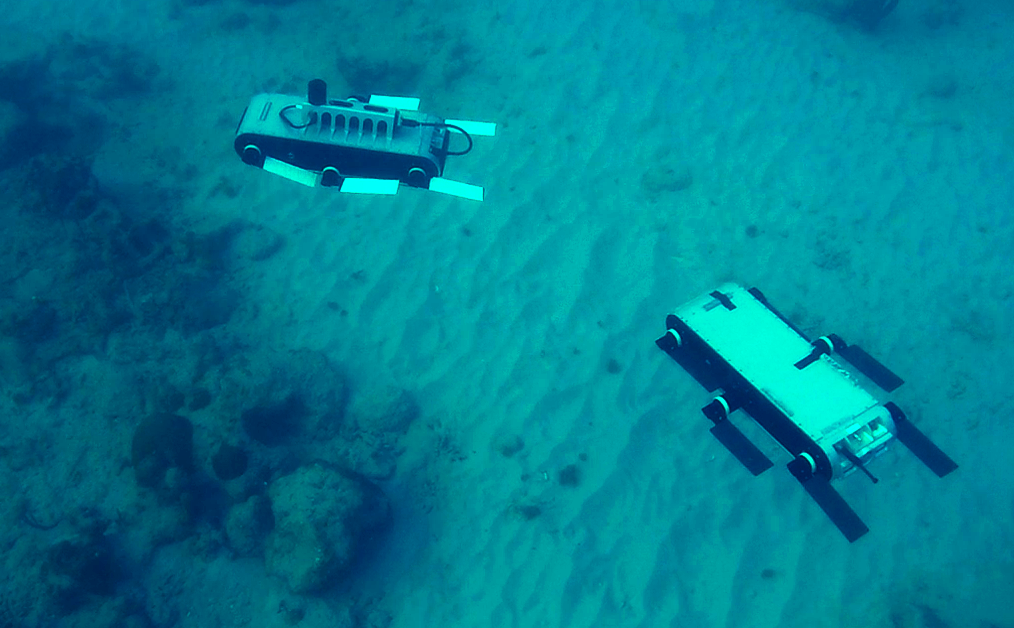}
\caption{A sample image from our underwater convoying field trial using Aqua hexapods~\cite{sattar2008enabling}. Videos of our field trials, datasets, code, as well as more information about the project are available at \small{\url{http://www.cim.mcgill.ca/~mrl/robot_tracking}}
}
\vspace{-0.5cm}
\end{figure}
While this strategy increases the complexity of the tracking task, it also offers the potential for greater robustness to changing pose variations of the target in which any attached markers may not be visible. Other works have demonstrated successful tracking methods using auxiliary devices for underwater localization, including mobile beacons~\cite{chandrasekhar2006localization}, aerial drones~\cite{erol2007auv}, or acoustic sampling~\cite{corke2007experiments}.
While these alternative strategies can potentially be deployed for multi-robot convoy tasks, they require additional costly hardware.

Our system learns visual features of the desired target from multiple views, through an annotated dataset of underwater video of the Aqua family of hexapod amphibious robots~\cite{sattar2008enabling}. This dataset is collected from both on-board cameras of a trailing robot as well as from diver-collected footage. Inspired by recent general-purpose object detection solutions, such as ~\cite{redmon2016yolo9000}, ~\cite{renNIPS15fasterrcnn}, ~\cite{squeezenet2016}, we propose several efficient neural network architectures for this specific robot tracking task, and compare their performance when applied to underwater sceneries. 

In particular, we compare methods using convolutional neural networks (CNNs), recurrent methods stacked on top of CNN-based methods, and frequency-based methods which track the gait frequency of the swimming robot. Furthermore, we demonstrate in an open-water field trial that one of our proposed architectures, based on YOLO~\cite{redmon2016you} and scaled down to run on-board the Aqua family of robots without GPU acceleration, is both efficient and does not sacrifice performance and robustness in the underwater robot-tracking domain despite motion blur, lighting variations and scale changes.

\section{RELATED WORK}
\subsection{Vision-Based Convoying}
Several vision-based approaches have shown promise for convoying in constrained settings. Some methods employ shared feature tracking to estimate all of the agents' positions along the relative trajectory of the convoy, with map-sharing between the agents. Avanzini \emph{et al.} demonstrate this with a SLAM-based approach~\cite{avanzini_slam_and_convoys}. However, these shared-feature methods require communication between the agents which is difficult without specialized equipment in underwater robots.

Using both visual feedback combined with explicit behavior cues to facilitate terrestrial robot convoys has also been considered~\cite{Dudek1995c}. Tracking was enhanced by both suitable engineered surface markings combined with action sequences that cue upcoming behaviors. Unlike the present work, that work was restricted to simple 2D motion and hand-crafted visual markings and tracking systems.

Other related works in vision-based convoying often employ template-based methods with fiducial markers placed on the leading agent \cite{schneiderman1995vision,fries2014monocular}. Such methods match the template to the image to calculate the estimated pose of the leading robot. While these methods could be used in our setting, we wish to avoid hand-crafted features or any external fiducial markers due to the possibility that these markers turn out of view of the tracking agent.

An example of a convoying method using visual features of the leading agent without templates or fiducial markers is \cite{giesbrecht2009vision}, which uses color-tracking mixed with SIFT features to detect a leading vehicle in a convoy. While we could attempt to employ such a method in an underwater scenario, color-based methods may not work as well due to the variations in lighting and color provided by underwater optics.



\subsection{Tracking Methods}
The extensive literature on visual tracking can be separated into \emph{model-based} and \emph{model-free} methods, each with their own set of advantages and drawbacks. 

In model-free tracking the algorithm has no prior information on the instance or class of objects it needs to track. Algorithms in this category, such as~\cite{cogd}, are typically initialized with a bounding box of an arbitrary target, and they adapt to viewpoint changes through semi-supervised learning. The TLD tracker \cite{kalal2012-tld}, for example, trains a detector online using positive and negative feedback obtained from image-based feature tracks. In general, tracking systems in this category suffer from \emph{tracking drift}, which is the accumulation of error over time either from false positive identification of unseen views of the target, or errors due to articulated motion, resulting in small accumulating errors leading to a drift away from the target object.

In model-based tracking, the algorithm is either trained on or has access to prior information on the appearance of the target. This can take the form of a detailed CAD model, such as in \cite{manz_vehicle_tracking}, which uses a 3D model describing the geometry of a car in order to improve tracking of the vehicle in image space. Typically, line and corner features are used in order to register the CAD model with the image. In our work we opted to avoid these methods because of their susceptibility to errors in terms of occlusion and non-rigid motion.

Works such as~\cite{bolme2010-corrfilter,nam2016mdnet} use convolutional neural networks and rely on supervised learning to learn a generic set of target representations. Our work herein is more closely related to this body of work, however, we are interested in a single target with known appearance. 




\section{DATASET COLLECTION} \label{dataset}
We collected a dataset\footnote{The dataset, along with its ground truth annotations, can be found at \url{http://www.cim.mcgill.ca/~mrl/robot_tracking}} of video recordings of the Aqua family of hexapods during field trials at McGill University's Bellairs Research Institute in Barbados. This dataset includes approximately 5200 
third-person view images extracted from video footage of the robots swimming in various underwater environments recorded from the diver's point of view and nearly 10000 
first-person point of view frames extracted from the tracking robot's front-facing camera.

We use the third-person video footage for training and validation, and the first-person video footage as a test set. This separation also highlights the way we envision our system to become widely applicable: the training data can be recorded with a handheld unit without necessitating footage from the robot's camera.

In the third-person video dataset the robot is usually maintained at the center of the image by the diver, which is not ideal for constructing a representative training set for neural networks, so we have augmented the dataset with synthetic images where the foreground image of the robot is overlaid on robot-free underwater backgrounds at different scales and translations over the image. This synthetic dataset contains about 48000 synthetic images, and is only used in the training of the VGG methods (as YOLO/Darknet performs its own data augmentation during training by default). In other words, we do not generate synthetic movie datasets for any of the recurrent methods that we examine here.
\section{DETECTION METHODS}

\subsection{Non-Recurrent Methods}

\subsubsection{VGG}
The VGG architecture~\cite{simonyan2014-VGG} classification performance has been shown to generalize well to several visual benchmark datasets in localization and classification tasks, so we use it as a starting point for tracking a single object. In particular we started from the VGG16 architecture, which consists of 16 layers, the first 13 of which are convolutional or max-pooling layers, while the rest are fully connected layers\footnote{Specifically, two fully connected layers of width 4096, followed by one fully connected layer of width 1000, denoted FC-4096 and FC-1000 respectively.}, the output of which is the classification or localization prediction of the network.

%
%
%
In our case, we want to output the vector $z=(x,y,w,h,p)$, where $(x,y)$ are the coordinates of the top left corner and $(w,h)$ is the width and height of the predicted bounding box. We normalize these coordinates to lie in $[0,1]$. $p$ is interpreted as the probability that the robot is present in the image. The error function that we want to minimize combines both the classification error, expressed as binary cross-entropy, and the regression error for localization, which in our case is the mean absolute error for true positives. More formally, the loss function that we used, shown here for a single data point, is:         
\begin{equation}
  L_n = \mathds{1}_{\bar{p}=1}\sum_{i=0}^3|z_i-\bar{z}_i|-\left(\bar{p}\text{log}(p) + (1-\bar{p})\text{log}(1-p) \right) \nonumber
\end{equation}
\noindent where symbols with bars denote ground truth annotations.

We evaluated the following variants of this architecture on our dataset:
\begin{itemize}
\item\small\emph{VGG16a}: the first 13 convolutional layers from VGG16, followed by two FC-128 ReLU, and a FC-5 sigmoid layer. We use batch normalization in this variant. The weights of all convolutional layers are kept fixed from pre-training.
\item\small\emph{VGG16b}: the first 13 convolutional layers from VGG16, followed by two FC-128 Parametric ReLU, and a FC-5 sigmoid layer. We use Euclidean weight regularization for the fully connected layers. The weights of all convolutional layers are kept fixed, except the top one.  
\item\small\emph{VGG16c}: the first 13 convolutional layers from VGG16, followed by two FC-228 ReLU, and a FC-5 sigmoid layer. The weights of all convolutional layers are fixed, except the top two.
\item\small\emph{VGG15}: the first 12 convolutional layers from VGG16, followed by two FC-128 ReLU, and a FC-5 sigmoid layer. We use batch normalization, as well as Euclidean weight regularization for the fully connected layers. The weights of all convolutional layers are kept fixed.
\item\small\emph{VGG8}: the first 8 convolutional layers from VGG16, followed by two FC-128 ReLU, and a FC-5 sigmoid layer. We use batch normalization in this variant, too. The weights of all convolutional layers are kept fixed. 
\end{itemize}

\noindent In all of our variants, we pre-train the network on the ImageNet dataset as in~\cite{simonyan2014-VGG} to drastically reduce training time and scale our dataset images to $(224,224,3)$ to match the ImageNet scaling.

\subsubsection{YOLO}
The YOLO detection system~\cite{redmon2016you} frames detection as a regression problem, using a single network optimized end-to-end to predict bounding box coordinates and object classes along with a confidence estimate. It enables faster predictions than most detection systems that are based on sliding window or region proposal approaches, while maintaining a relatively high level of accuracy.

We started with the TinyYOLOv2 architecture~\cite{redmon2016yolo9000}, but we found that inference was on our robot's embedded platform (without GPU acceleration) was not efficient enough for fast, closed-loop, vision-based, onboard control. Inspired by lightweight architectures such as~\cite{squeezenet2016}, we condensed the TinyYOLOv2 architecture as shown in Table \ref{table:yolov2_arch}. This enabled inference on embedded robot platforms at reasonable frame rates (13 fps). Following Ning~\footnote{'YOLO CPU Running Time Reduction: Basic Knowledge and Strategies' at \url{https://goo.gl/xaUWjL}} and \cite{squeezenet2016} we:  

%

\begin{itemize}
\item replace some of the $3\times 3$ filters with $1\times 1$ filters, and
\item decrease the depth of the input volume to $3\times 3$ filters.
\end{itemize}

\noindent Our ReducedYOLO architecture is described in Table~\ref{table:squeezeyolov2_arch}.
This architecture keeps the same input resolution and approximately the same number of layers in the network, yet drastically decreases the number of filters for each layer. Since we started with a network which was designed for detection tasks of up to 9000 classes in the case of TinyYOLOv2 \cite{redmon2016yolo9000}, we hypothesize that the reduced capacity of the network would not significantly hurt the tracking performance for a single object class. This is supported by our experimental results. Additionally, we use structures of two $1\times 1$ filters followed by a single $3 \times 3$ filter, similar to Squeeze layers in SqueezeNet \cite{squeezenet2016}, to compress the inputs to $3\times 3$ filters. Similarly to VGG \cite{simonyan2014-VGG} and the original YOLO architecture \cite{redmon2016you}, we double the number of filters after every pooling step.

As in the original TinyYOLOv2 configuration, both models employ batch normalization and leaky rectified linear unit activation functions on all convolutional layers.

\begin{table}[h!]
\centering
 \begin{tabular}{c | c | c | c} 
 Type & Filters & Size/Stride & Output \\ [0.5ex] 
 \hline\hline
 Input &  &  & $416\times 416$  \\ 
 Convolutional & 16 & $3\times 3$/1 & $416\times 416$  \\ 
 Maxpool &  & $2\times 2$/2 & $208\times208$ \\ 
 Convolutional & 32 & $3\times 3$/1 &  $208\times208$ \\ 
 Maxpool &  & $2\times 2$/2 & $104\times104$ \\ 
 Convolutional & 64 & $3\times 3$/1 & $104\times104$  \\ 
 Maxpool &  & $2\times 2$/2 & $52\times52$ \\ 
 Convolutional & 128 & $3\times 3$/1 & $52\times52$  \\ 
 Maxpool &  & $2\times 2$/2 & $26\times26$ \\ 
 Convolutional & 256 & $3\times 3$/1 &  $26\times26$ \\ 
 Maxpool &  & $2\times 2$/2 & $13\times13$ \\ 
 Convolutional & 512 & $3\times 3$/1 &  $13\times13$ \\ 
 Maxpool &  & $2\times 2$/1 & $13\times13$ \\ 
 Convolutional & 1024 & $3\times 3$/1 &  $13\times13$ \\ 
 Convolutional & 1024 & $3\times 3$/1 &  $13\times13$ \\ 
 Convolutional & 30 & $1\times 1$/1 &  $13\times13$ \\ 
 \hline \hline
 Detection &  &  &   \\ 
 \end{tabular}
 \caption{TinyYOLOv2 architecture}
 \label{table:yolov2_arch}
\end{table}

\begin{table}[h!]
\centering
 \begin{tabular}{c | c | c | c} 
 Type & Filters & Size/Stride & Output \\ [0.5ex] 
 \hline\hline 
 Input &  &  & $416\times 416$  \\ 
 Convolutional & 16 & $7\times 7$/2 &  $208\times208$ \\ 
 Maxpool &  & $4\times 4$/4 & $52\times52$ \\ 
 Convolutional & 4 & $1\times 1$/1 & $52\times52$  \\ 
 Convolutional & 4 & $1\times 1$/1 & $52\times52$  \\ 
 Convolutional & 8 & $3\times 3$/1 &  $52\times52$ \\ 
 Maxpool &  & $2\times 2$/2 & $26\times26$ \\ 
 Convolutional & 8 & $1\times 1$/1 & $26\times26$  \\ 
 Convolutional & 8 & $1\times 1$/1 & $26\times26$  \\ 
 Convolutional & 16 & $3\times 3$/1 &  $26\times26$ \\ 
 Maxpool &  & $2\times 2$/2 & $13\times13$ \\ 
 Convolutional & 32 & $3\times 3$/1 &  $13\times13$ \\ 
 Maxpool &  & $2\times 2$/2 & $6\times6$ \\ 
 Convolutional & 64 & $3\times 3$/1 &  $6\times6$ \\ 
 Convolutional & 30 & $1\times 1$/1 &  $6\times6$ \\ 
 \hline \hline
 Detection &  &  &   \\ 
 \end{tabular}
 \caption{Our ReducedYOLO architecture}
 \label{table:squeezeyolov2_arch}
\end{table}

\subsection{Recurrent Methods}
In vision-based convoying, the system may lose sight of the object momentarily due to occlusion or lighting changes, and thus lose track of its leading agent. In an attempt to address this problem, we use recurrent layers stacked on top of our ReducedYOLO architecture, similarly to~\cite{ning2016spatially}. In their work, Ning~\emph{et al.} use the last layer of features output by the YOLO network for $n$ frames (concatenated with the YOLO bounding box prediction which has the highest IOU with the ground truth) and feed them to single forward Long-Term Short-Term Memory Network (LSTM).

While Ning \textit{et al.} assume that objects of interest are always in the image (as they test on the OTB-100 tracking dataset), we instead assume that the object may not be in frame. Thus, we make several architectural modifications to improve on their work and make it suitable for our purposes. First, Ning \textit{et al.} use a simple mean squared error (MSE) loss between the output bounding box coordinates and the ground truth in addition to a penalty which minimizes the MSE between the feature vector output by the recurrent layers and the feature vector output of the YOLO layers. We find that in a scenario where there can be images with no bounding box predicted (as is the case in our system), this makes for an extremely unstable objective function. Therefore we instead use a modified YOLO objective for our single-bounding box single class case. This results in Recurrent ReducedYOLO (RROLO) having the following objective function, shown here for a single data point:
\begin{multline}
I((\sqrt{\bar{x}} - \sqrt{x})^2 + (\sqrt{\bar{y}} - \sqrt{y})^2)\alpha_{coord}\\
+I((\sqrt{\bar{w}} - \sqrt{w})^2 + (\sqrt{\bar{h}} - \sqrt{h})^2)\alpha_{coord}\\
+ I(\text{IOU} - p)^2 \alpha_{obj}
\\+ (1-I)(\text{IOU} - p)^2 \alpha_{no\_obj}
\end{multline}
\noindent where $\alpha_{coord}, \alpha_{obj}, \alpha_{no\_obj}$ are tunable hyper-parameters (left at 5, 1, 0.5 respectively based on the original YOLO objective), $\bar{w}, \bar{h}, w, h$ are the width, height, predicted width and predicted height, respectively, $I \in \{0,1\}$ indicates whether the object exists in the image according to ground truth, $p$ is the confidence value of the prediction and IOU is the Intersection Over Union of the predicted bounding box with the ground truth.

Furthermore, to select which bounding box prediction of YOLO to use as input to our LSTM (in addition to features), we use the highest confidence bounding box rather than the one which overlaps the most with the ground truth. We find that the latter case is not a fair comparison or even possible for real-world use and thus eliminate this assumption.

In order to drive the final output to a normalized space (ranging from 0 to 1), we add fully connected layers with sigmoidal activation functions on top of the final LSTM output, similarly to YOLOv2~\cite{redmon2016yolo9000}. Redmon and Farhadi posit that this helps stabilize the training due to the normalization of the gradients at this layer. We choose three fully connected layers with $|\text{YOLO}_{output}|, 256, 32$ hidden units (respectively) and a final output of size $5$. We also apply dropout on the final dense layers at training time with a probability of $.6$ that the weight is kept.

\begin{figure}[hb!]
  \centering
  \includegraphics[width=8cm]{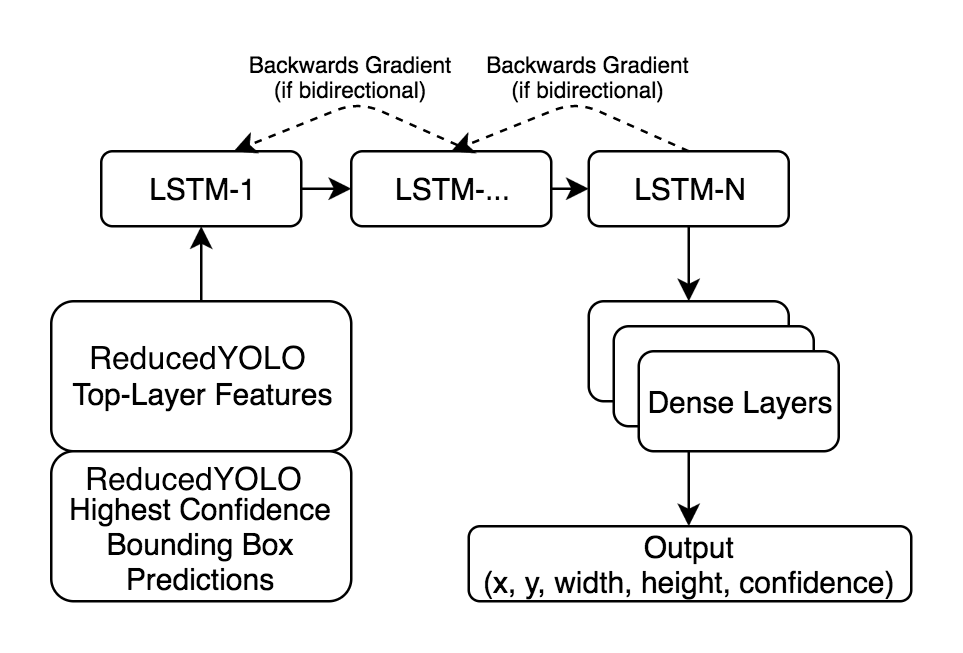}
  \caption{Overview of our Recurrent ReducedYOLO (RROLO) architecture. The original ROLO work~\cite{ning2016spatially} did not use bidirectional, dense layers, or multiple LSTM cells in their experiments.}
  \label{fig:lstm}
\end{figure}

We also include multi-layer LSTMs to our experimental evaluation as well as bidirectional LSTMs which have been shown to perform better on longer sequences of data \cite{graves2012supervised}. A general diagram of our LSTM architecture can be seen in Figure~\ref{fig:lstm}. 
Our recurrent detection implementation, based partially on code provided by~\cite{ning2016spatially}, is made publicly accessible.\footnote{\url{https://github.com/Breakend/TemporalYolo}}

\subsection{Methods Based on Frequency-Domain Detection}
Periodic motions have distinct frequency-domain signatures that can be used as reliable and robust features for visual detection and tracking. Such features have been used effectively \cite{sattar2009underwater, islam2017mixed} by underwater robots to track scuba divers. Flippers of a human diver typically oscillate at frequencies between $1$ and $2$ Hz, which produces periodic intensity variations in the image-space over time. These variations correspond to distinct signatures in the frequency-domain (high-amplitude spectra at $1$-$2$Hz), which can be used for reliable detection. While for convoying purposes, the lead robot's flippers may not have such smoothly periodic oscillations, the frequency of the flippers is a configurable parameter which would be known beforehand. 

\begin{figure}[ht]
\centering
\includegraphics [width=\linewidth]{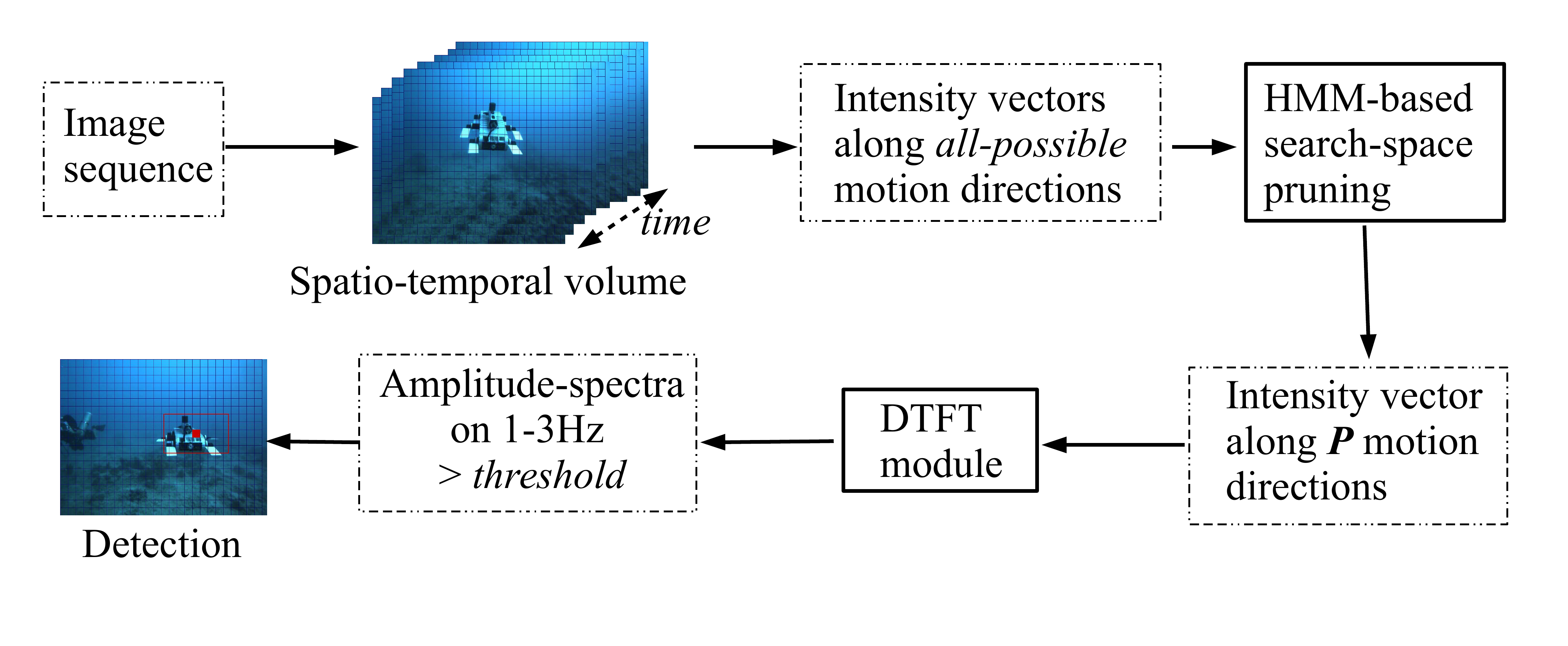}
\vspace*{-10mm}
\caption{Outline of mixed-domain periodic motion (MDPM) tracker \cite{islam2017mixed}}
\label{fig:FTracker}
\vspace{-0.3cm}
\end{figure}

We implement the mixed-domain motion (MDPM) tracker described by Islam {\em et al}~\cite{islam2017mixed}. An improved version of Sattar {\em et al}~\cite{sattar2009underwater}, the MDPM works as follows (illustrated in Figure \ref{fig:FTracker}) : 
\begin{itemize}
\item First, intensity values are captured along arbitrary motion directions; motion directions are modeled as sequences of non-overlapping image sub-windows over time.
          
\item By exploiting the captured intensity values, a Hidden Markov Model (HMM)-based pruning method discards motion directions that are unlikely to be directions where the robot is swimming.

\item A  Discrete Time Fourier Transform (DTFT) converts the intensity values along $P$ most potential motion directions to frequency-domain amplitude values. High amplitude spectra on $1$-$3$Hz is an indicator of robot motion, which is subsequently used to locate the robot in the image space.

\end{itemize}

\section{VISUAL SERVOING CONTROLLER}
The Aqua family of underwater robots allows 5 degrees-of-freedom\footnote{Yaw, pitch, and roll rate, as well as forward and vertical speed} control, which enables agile and fast motion in 3D. This characteristic makes vehicles of this family ideal for use in tracking applications that involve following other robots as well as divers~\cite{sattar2009underwater}. 

One desired attribute of controllers in this type of setting is that the robot moves smoothly enough to avoid motion blur, which would degrade the quality of visual feedback. To this end we have opted for an image-based visual servoing controller that avoids explicitly estimating the 3D position of the target robot in the follower's camera coordinates, as this estimate typically suffers from high variance along the optical axis. This is of particular relevance in the underwater domain because performing camera calibration underwater is a time-consuming and error-prone operation. Conversely, our tracking-by-detection method and visual servoing controller do not require camera calibration. 
\begin{figure}[ht]
\centering
\includegraphics [width=0.65\linewidth]{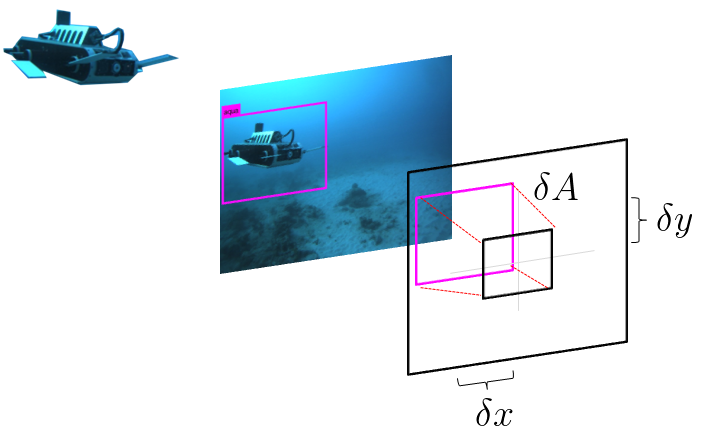}
\caption{Errors used by the robot's feedback controller. $\delta x$ is used for yaw control, $\delta y$ for depth control, and the error in bounding box area, $\delta A$ is used for forward speed control.}
\label{fig:tracking_errors}
\end{figure}
Our controller regulates the motion of the vehicle to bring the observed bounding boxes of the target robot on the center of the follower's image, and also to occupy a desired fraction of the total area of the image. It uses a set of three error sources, as shown in Fig.~\ref{fig:tracking_errors}, namely the 2D translation error from the image center, and the difference between the desired and the observed bounding box area. 

The desired roll rate and vertical speed are set to zero and are handled by the robot's 3D autopilot~\cite{iros2014_megerShkurtiCortesPozaGiguereDudek}. The translation error on the x-axis, $\delta x$, is converted to a yaw rate through a PID controller. Similarly, the translation error on the y-axis, $\delta y$, is scaled to a desired depth change in 3D space. When the area of the observed bounding box is bigger than desired, the robot's forward velocity is set to zero. We do not do a backup maneuver in this case, even though the robot supports it, because rotating the legs $180^o$ is not an instantaneous motion. The difference in area of the observed versus the desired bounding box, namely $\delta A$, is scaled to a forward speed. Our controller sends commands  at a rate of 10Hz and assumes that a bounding box is detected at least every 2 seconds, otherwise it stops the robot.    

\section{EXPERIMENTAL RESULTS}
We evaluate each of the implemented methods on the common test dataset described in section \ref{dataset} using the metrics described below, with $n_{images}$ the total number of test images, $n_{TP}$ the number of true positives, $n_{TN}$ the number of true negatives, $n_{FN}$ the number of false negatives and $n_{FP}$ the number of false positives: 
\begin{itemize}
\item Accuracy : $\frac{n_{TP} + n_{TN}}{n_{images}}$ 
\item Precision : $\frac{n_{TP}}{n_{TP} + n_{FP}}$ and recall: $\frac{n_{TP}}{n_{TP} + n_{FN}}$
\item Average Intersection Over Union (IOU) : Computed from the predicted and ground-truth bounding boxes over all true positive detections (between 0 and 1, with 1 being perfect alignment)
\item Localization failure rate (LFR): Percentage of true positive detections having IOU under 0.5  \cite{cehovin-vot15metrics}
\item Frames per second (FPS) : Number of images processed/second
\end{itemize}
Each of the implemented methods outputs its confidence that the target is visible in the image. We chose this threshold for each method by generating a precision-recall curve and choosing the confidence bound which provides the best recall tradeoff for more than $95\%$ precision.


\begin{table}[h!]
\centering
 \begin{tabular}{|c c c c c c c|} 
 \hline
 Algorithm & ACC & IOU & P & R & FPS & LFR\\ [0.5ex] 
 \hline\hline
 VGG16a &0.80&\textbf{0.47}&0.78&\textbf{0.79}&1.68&\textbf{47\%} \\ [1ex]
 VGG16b &\textbf{0.82}&0.39&0.85&0.73&1.68&61\% \\ [1ex]
 VGG16c &0.80&0.35&\textbf{0.88}&0.64&1.68&70\% \\ [1ex]
 VGG15 &0.71&0.38&0.65&0.75&1.83&62\% \\ [1ex]
 VGG8 &0.61&0.35&0.55&0.69&\textbf{2.58}&65\% \\ [1ex]
 \hline
 TinyYOLOv2 & \textbf{0.86} & \textbf{0.54} & \textbf{0.96} & \textbf{0.88} & 0.91 & \textbf{34}\%\\ 
 ReducedYOLO & 0.85 & 0.50 & 0.95 & 0.86 & \textbf{13.3} & 40\%\\ 
 \hline
  RROLO (n=3,z=1) & 0.68 & 0.53 & 0.95 & 0.64 & \textbf{12.69} & 15\%\\ 
  RROLO (n=3,z=2) & \textbf{0.84} & \textbf{0.54} & \textbf{0.96} & \textbf{0.83} & 12.1 & \textbf{9}\%\\ 
  RROLO (n=6,z=1) & 0.81 & 0.53 & 0.95 & 0.80 & 12.1 & 11\%\\ 
  RROLO (n=6,z=2) & 0.83 & \textbf{0.54} & \textbf{0.96} & 0.82 & 12.03 & 11\%\\ 
  RROLO (n=9,z=1) & 0.81 & 0.53 & 0.95 & 0.80 & 11.53 & \textbf{9\%}\\ 
  RROLO (n=9,z=2) & 0.80 & 0.50 & \textbf{0.96} & 0.79 & 11.36 & 14\%\\ 
 \hline
 MDPM Tracker & 0.25 & \textbf{0.16} & 0.94 & 0.26 & \textbf{142} & \textbf{19.3\%} \\
 TLD Tracker & \textbf{0.57} & 0.12 & \textbf{1.00} & \textbf{0.47} & 66.04 & 97\% \\
 \hline
 \end{tabular}
 \caption{Comparison of all tracking methods. Precision and recall values based on an optimal confidence threshold.}
 \label{table:results}
 \vspace{-0.3cm}
\end{table}

We present the evaluation results in Table \ref{table:results} for each of the algorithms that we considered. 
The \textit{FPS} metric was measured across five runs on a CPU-only machine with a 2.7GHz Intel i7 processor. 

\subsection{Non-Recurrent Methods}
As we can see in Table \ref{table:results}, the original TinyYOLOv2 model is the best performing method in terms of IOU, precision, and failure rate.

However our results show that the ReducedYOLO model achieves a 14x speedup over TinyYOLOv2,  without significantly sacrificing accuracy. This is a noteworthy observation since ReducedYOLO uses 3.5 times fewer parameters compared to TinyYOLOv2. This speedup is crucial for making the system usable on mobile robotic platforms which often lack Graphical Processing Units, and are equipped with low-power processing units. Additionally, note that the precision metric has not suffered while reducing the model, which implies that the model rarely commits false positive errors, an important quality in convoying where a single misdetection could deviate the vehicle off course. 

In addition, we found that none of the VGG variants fared as well as the YOLO variants, neither in terms of accuracy nor in terms of efficiency. The localization failure rate of the VGG variants was reduced with the use of batch normalization. Increasing the width of the fully connected layers and imposing regularization penalties on their weights and biases did not lead to an improvement over \emph{VGG16a}. Reducing the total number of convolutional layers, resulting in the \emph{VGG8} model lead to a drastic decrease in both classification and localization performance, which suggests that even when trying to detect a single object, network depth is necessary for VGG-type architectures.      

Finally, the TLD tracker~\cite{kalal2012-tld} performed significantly worse than any of the detection-based methods, mainly due to tracking drift. It is worth noting that we did not reinitialize TLD after the leading robot exited the field of view of the following robot, and TLD could not always recover. This illustrates why model-free trackers are in general less suitable for convoying tasks than model-based trackers.

\subsection{Recurrent Methods}
All the recurrent methods were trained using features obtained from the ReducedYOLO model, precomputed on our training set. We limit our analysis to the recurrent model using the ReducedYOLO model's features, which we'll refer to as Recurrent ReducedYOLO (RROLO), since this model can run closest to real time on our embedded robot system. Our results on the test set are shown in Table~\ref{table:results}. For these methods, $z$ denotes the number of LSTM layers, $n$ is the number of frames in a given sequence. Note that the runtime presented here for RROLO methods includes the ReducedYOLO inference time. Finally, while bidirectional recurrent architectures were implemented and tested as well, we exclude results from those models in the table as we found that in our case these architectures resulted in worse performance overall across all experiments. 

As can be seen in Table~\ref{table:results}, we find that the failure rate and predicted confidence can be tuned and improved significantly without impacting precision, recall, accuracy, or IOU. More importantly, we find that the correlation between bounding box IOU (with the ground truth) and the predicted confidence value of our recurrent methods is much greater than any of the other methods, which translates to a more interpretable model with respect to the confidence threshold parameter while also reducing the tracking failure rate. For our best configurations of VGG, YOLO, ReducedYOLO, and RROLO, we take the Pearson correlation r-value and the mean absolute difference between the ground truth IOU and the predicted confidence\footnote{Pearson correlation r-value: VGG (.70), YOLO (.48), ReducedYOLO (.56), RROLO (.88). Mean absolute difference between confidence predicted and IOU with ground truth: VGG (.37), YOLO (.17), ReducedYOLO (.18), RROLO (.08).}. We find that RROLO overall is the most correlated and has the least absolute difference between the predicted confidence and ground truth IOU. 


Variations in layers and timesteps do not present a significant difference in performance, while yielding a significant reduction in failure rate even with short frame sequences ($n=3,z=2$). Furthermore, the frame rate impact is negligible with a single layer LSTM and short time frames, so we posit that it is only beneficial to use a recurrent layer on top of ReducedYOLO. 
While the best length of the frame sequence to examine may vary based on characteristics of the dataset, in our case $n=3, z=2$ provides the best balance of speed, accuracy, IOU, precision and recall, since this model boosts all of the evaluation metrics while retaining IOU with the ground truth and keeping a relatively high FPS value.

Increasing the number of LSTM layers can boost accuracy and recall further, without impacting IOU or precision significantly, at the expense of higher runtime and higher risk of overfitting. We attempted re-balancing and re-weighting the objective in our experiments and found that the presented settings worked best. We suspect that no increase in IOU, precision and recall was observed as there may not be enough information in the fixed last layers of the YOLO output to improve prediction. Future work to improve the recurrent system would target end-to-end experiments on both the convolutional and recurrent layers, along with experiments investigating different objective functions to boost the IOU while making the confidence even more correlated to IOU.

%
%
%
%
%
\subsection{Method Based on Frequency-Domain Detection}
In our implementation of MDPM tracker, non-overlapping sub-windows of size $30\times30$ pixels over $10$ sequential frames are considered to infer periodic motion of the robot. Peaks in the amplitude spectrum in the range $1$-$3$Hz constitute an indicator of the robot's direction of motion. 
We found that the frequency responses generated by the robot's flippers are not strong and regular. This is due to lack of regularity and periodicity in the robot's flipping pattern (compared to that of human divers), but also due to the small size of the flippers compared to the image size. Consequently, as Table \ref{table:results} suggests, MDPM tracker exhibits poor performance in terms of accuracy, recall, and IOU. The presence of high amplitude spectra at $1$-$3$Hz indicates the robot's motion direction with high precision. However, these responses are not regular enough and therefore the algorithm fails to detect the robot's presence in a significant number of detection cycles. We can see however that the failure rate for this method is one of the lowest among the studied methods, indicating very precise bounding boxes when detections do occur. Additionally, this method does not need training and is the fastest (and least computationally expensive) method, by a significant margin. Therefore given more consistent periodic gait patterns it would perform quite well, as previously demonstrated in~\cite{islam2017mixed}.

\subsection{Field Trial}
\subsubsection{Setup}
To demonstrate the practicality of our vision-based tracking system, we conducted a set of in-ocean robot convoying field runs by deploying two Aqua robots at 5 meters depth from the sea surface.
The appearance of the leading robot was altered compared to images in our training dataset, due to the presence of an additional sensor pack on its top plate. This modification allowed us to verify the general robustness of our tracking-by-detection solution, and specifically to evaluate the possibility of overfitting to our training environments.

\begin{figure}[ht!]
\centering
\includegraphics [width=\linewidth]{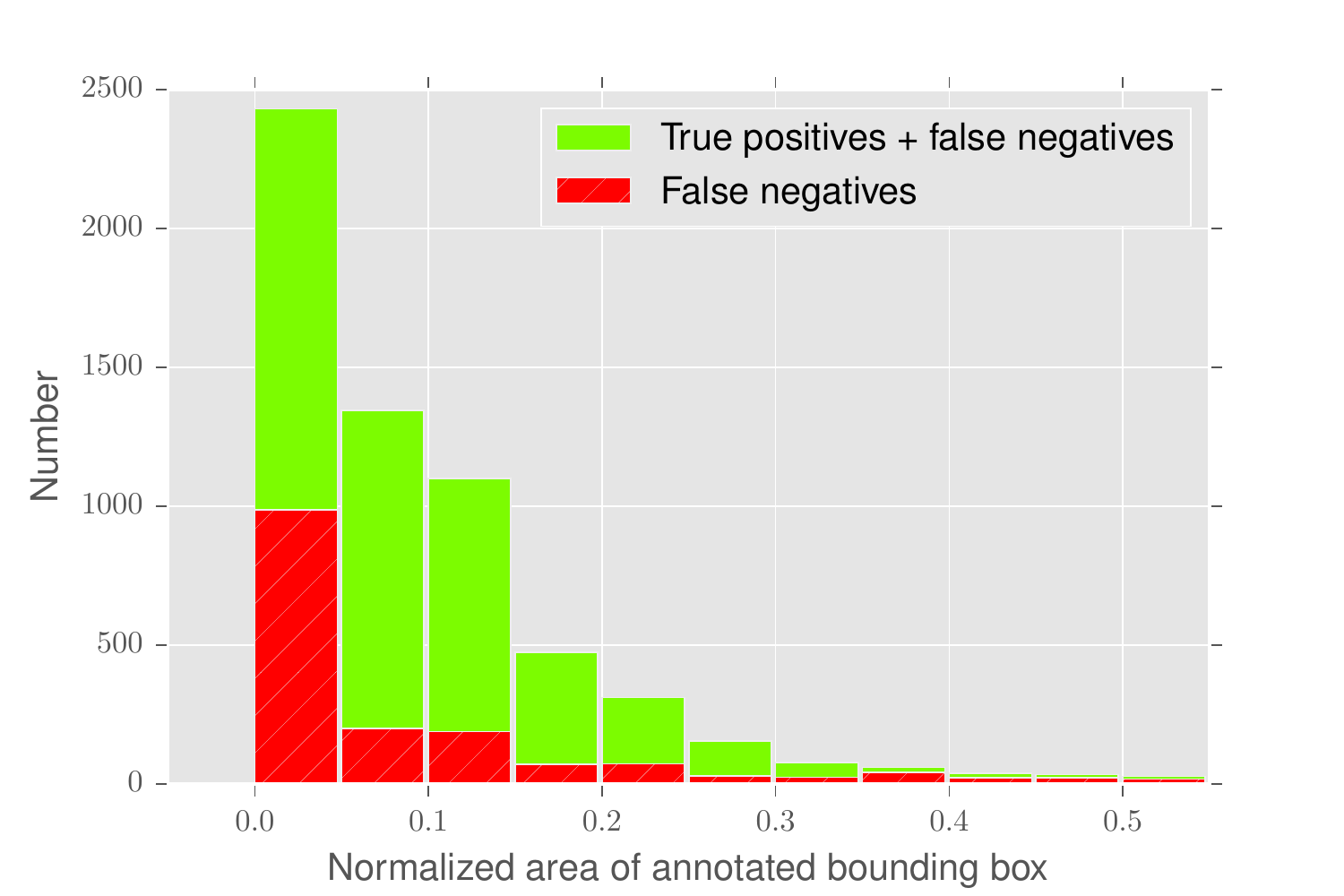}
\caption{Histogram of true positive and false negative detections as a function of the area of annotated bounding boxes, as obtained from in-ocean robot tracking runs.}
\label{fig:bbox_area_vs_tp_det_ratio}
\vspace{-0.4cm}
\end{figure}

We programmed the target robot to continuously execute a range of scripted trajectories, including maneuvers such as in-place turning, changing depth, and swimming forward at constant speed. We deployed the ReducedYOLO model on the follower robot, which operated at 7\,Hz onboard an Intel NUC i3 processor without GPU acceleration or external data tethering. Moreover, the swimming speeds of both robots were set to be identical in each run ($0.5-0.7\,m/s$), and they were initialized at approximately two meters away from one another, but due to currents and other factors the distance between them (and the scale of observed bounding boxes) changed throughout the experiment runs. 

\subsubsection{Results}

We configured the follower robot to try to track the leading robot at a fixed nominal distance. This was achieved by setting the desired bounding box area to be $50\%$ of the total image area, as seen in Fig.~\ref{fig:tracking_errors}. 

\begin{figure}[ht!]
\centering
\includegraphics[width=\linewidth]{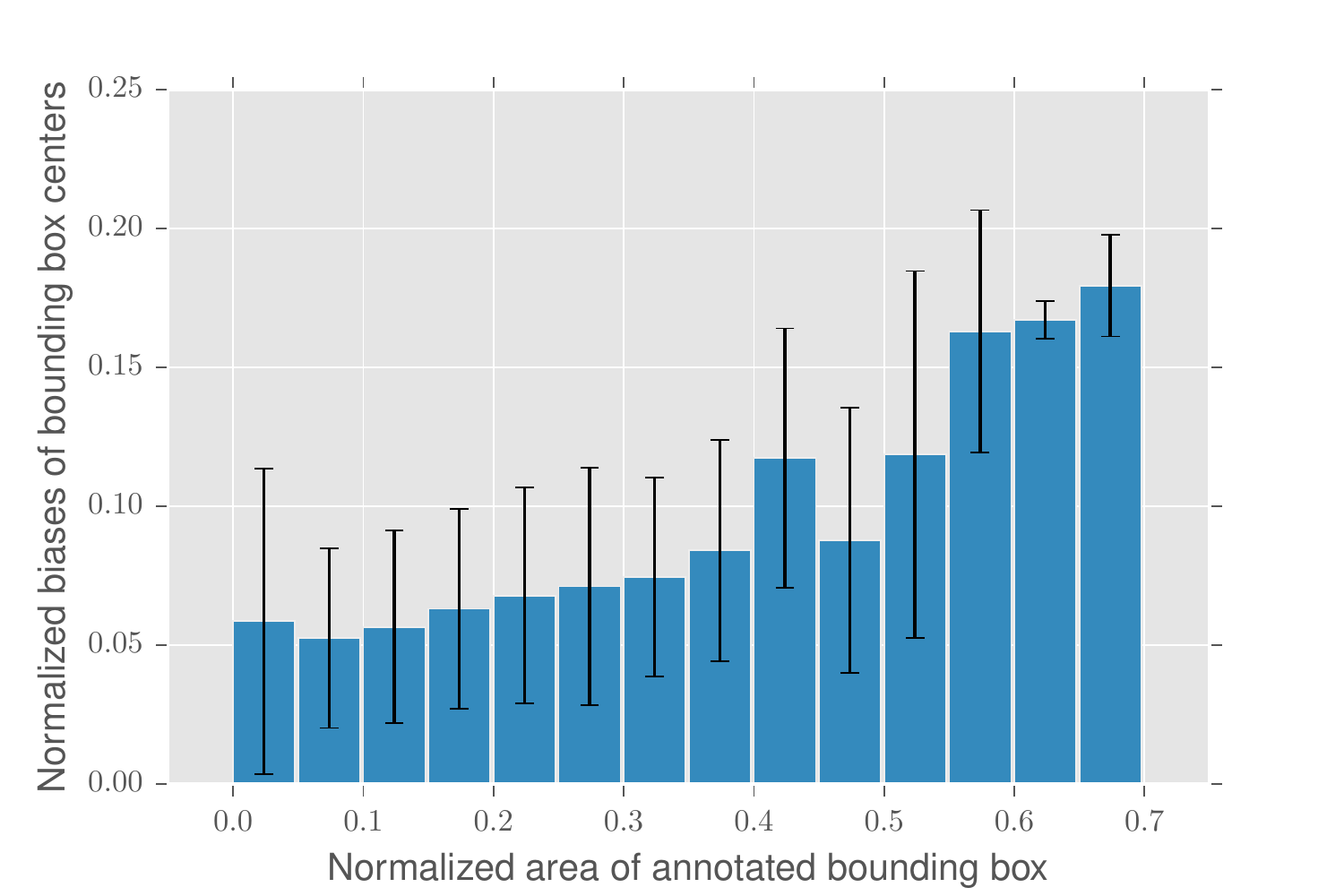}
\caption{Histogram of average biases between detected vs. annotated bounding box centers, as obtained from in-ocean robot tracking runs. Bars indicate 1$\sigma$ error.}
\label{fig:bbox_area_vs_tp_avg_mid_dev}
\vspace{-0.3cm}
\end{figure}

The ReducedYOLO detector consistently over-estimated the small size of the target.
Nevertheless, Fig.~\ref{fig:bbox_area_vs_tp_avg_mid_dev} indicates that the bias error in bounding box centers between detected versus ground truth was consistently low in each frame, regardless of the target size, on average within $10\%$ of the image's width to each other. This is notable due to frequent occurrences where the robot's size occupied less than $50\%$ of the total area of the image.

\begin{figure}[ht!]
\centering
\includegraphics[width=\linewidth]{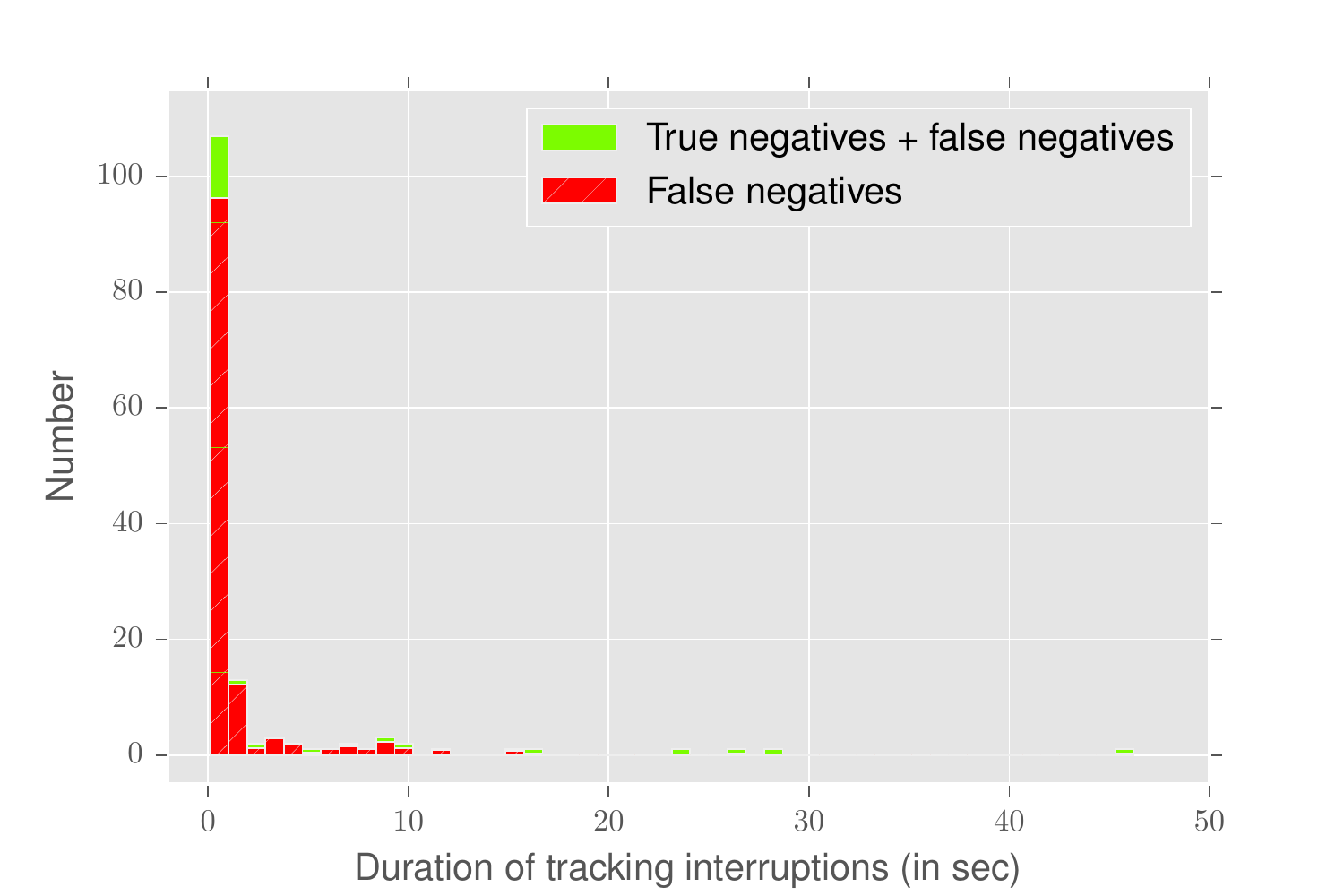}
\caption{Histogram of true negative and false negative classifications in terms of their duration for our ReducedYOLO model, as obtained from in-ocean robot tracking runs.}
\label{fig:false_negatives_vs_all_negatives}
\vspace{-0.4cm}
\end{figure}

We also evaluated the performance of our system in terms of the average \enquote{track length}, defined as the length of a sequence of true positive detections with a maximum of 3 seconds of interruption. Across all field trial runs, the follower achieved $27$ total tracks, with an average duration of $18.2$\,sec ($\sigma = 21.9$\,sec) and a maximum duration of $85$\,sec. As shown in Fig.~\ref{fig:false_negatives_vs_all_negatives}, the vast majority of tracking interruptions were short, specifically less than a second, which did not affect the tracking performance, as the leading robot was re-detected. The majority of these tracking interruptions were due to the fact that the annotated bounding box area of the leading robot was less than $20\%$ of the total area of the follower's image.    
%
%
Sustained visual detection of the target, despite significant visual noise and external forces in unconstrained natural environments, and without the use of engineered markers, reflects successful and robust tracking performance in the field.

\section{CONCLUSIONS}
We presented a general \emph{tracking-by-detection} approach that relies on visual sensing of the appearance of non-engineered targets, and also capitalizes on recent advances in deep learning for deployment onboard robots with limited computing capabilities.
We demonstrated the utility of several lightweight neural network architecture for appearance-based visual convoying, and we showed improvements made possible by recurrent extensions. We successfully performed multi-robot convoying in the open sea in practice, using supervised learning based on limited training data annotated beforehand. Furthermore, we carried out an extensive comparison of various tracker variants with respect to a multitude of desirable attributes for visual convoying tasks. 

In the future, we seek to improve on temporal-based bounding box detection by making the entire architecture trainable end-to-end. We also aim to extend this work to visual servoing with multiple bounding boxes per frame. We are working towards robust target tracking, despite interruptions in visual contact by using stronger predictive models.

\addtolength{\textheight}{-0cm}   





\section*{ACKNOWLEDGMENT}
\noindent This work was funded by the NSERC Canadian Field Robotics Network (NCFRN).

\bibliography{bibtex/bib/IEEEabrv.bib,bibtex/bib/IEEEexample.bib}{}
\bibliographystyle{IEEEtran}

\end{document}